\documentclass[conference]{IEEEtran}
\IEEEoverridecommandlockouts
\usepackage{cite}
\usepackage{amsmath,amssymb,amsfonts}
\usepackage{algorithmic}
\usepackage{graphicx}
\usepackage{textcomp}
\usepackage{xcolor}
\usepackage{cases}
\usepackage{url}
\usepackage[caption=false]{subfig}
\def\BibTeX{{\rm B\kern-.05em{\sc i\kern-.025em b}\kern-.08em
    T\kern-.1667em\lower.7ex\hbox{E}\kern-.125emX}}
\begin{document}

\makeatletter
\newcommand{\linebreakand}{%
  \end{@IEEEauthorhalign}
  \hfill\mbox{}\par
  \mbox{}\hfill\begin{@IEEEauthorhalign}
}
\makeatother

\title{Trade-off on Sim2Real Learning: Real-world Learning Faster than Simulations
}

\author{\IEEEauthorblockN{1\textsuperscript{st} Jingyi Huang}
\IEEEauthorblockA{\textit{School of Information Science} \\
\textit{and Technology}\\
\textit{ShanghaiTech University}\\
Shanghai, China \\
huangjy@shanghaitech.edu.cn}
\and
\IEEEauthorblockN{2\textsuperscript{nd} Yizheng Zhang}
\IEEEauthorblockA{\textit{School of Information Science} \\
\textit{and Technology}\\
\textit{ShanghaiTech University}\\
Shanghai, China \\
zhangyzh1@shanghaitech.edu.cn}
\and
\IEEEauthorblockN{3\textsuperscript{rd} Fabio Giardina}
\IEEEauthorblockA{\textit{John A. Paulson School of}\\
\textit{Engineering and Applied Sciences} \\
\textit{Harvard University}\\
Cambridge, Massachusetts, United States \\
giardina.fa@gmail.com}
\linebreakand 
\IEEEauthorblockN{4\textsuperscript{th} Andre Rosendo}
\IEEEauthorblockA{\textit{School of Information Science} \\
\textit{and Technology}\\
\textit{ShanghaiTech University}\\
Shanghai, China \\
arosendo@shanghaitech.edu.cn}
}

\maketitle

\begin{abstract}
Deep Reinforcement Learning (DRL) experiments are commonly performed in simulated environments due to the tremendous training sample demands from deep neural networks. In contrast, model-based Bayesian Learning allows a robot to learn good policies within a few trials in the real world. Although it takes fewer iterations, Bayesian methods pay a relatively higher computational cost per trial, and the advantage of such methods is strongly tied to dimensionality and noise. In here, we compare a Deep Bayesian Learning algorithm with a model-free DRL algorithm while analyzing our results collected from both simulations and real-world experiments. While considering Sim and Real learning, our experiments show that the sample-efficient Deep Bayesian RL performance is better than DRL even when computation time (as opposed to number of iterations) is taken in consideration. Additionally, the difference in computation time between Deep Bayesian RL performed in simulation and in experiments point to a viable path to traverse the reality gap. We also show that a mix between Sim and Real does not outperform a purely Real approach, pointing to the possibility that reality can provide the best prior knowledge to a Bayesian Learning. Roboticists design and build robots every day, and our results show that a higher learning efficiency in the real-world will shorten the time between design and deployment by skipping simulations.
\end{abstract}


\section{Introduction}

Reinforcement learning (RL) allows robots to adaptively learn an optimal behavior for specific tasks by a series of trial-and-error. A significant portion of RL algorithms follows the model-free paradigm and train on samples without emulating a transition model. These algorithms usually require many trials to learn a specific task. As a consequence, most of these applications are first performed in a simulated environment and then transferred to the real world. This transfer is very challenging due to the systematic difference between the simulator and the real environment, commonly known as the reality gap \cite{jakobi1995noise}. Deep Q Learning \cite{DQN2016} is a notable example from the model-free branch of algorithms. These algorithms have been commonly used in robotic control and decision-making since \cite{SurveyPolicy2019,polydoros2017surveymb} proposed the framework. DDPG \cite{ContiCtrl2016} and NAF \cite{ContiCtrlModel2016} adapted the ideas of DQL to the continuous control domain on simulations.

\begin{figure}
    \centering
    \includegraphics[width=0.9\columnwidth]{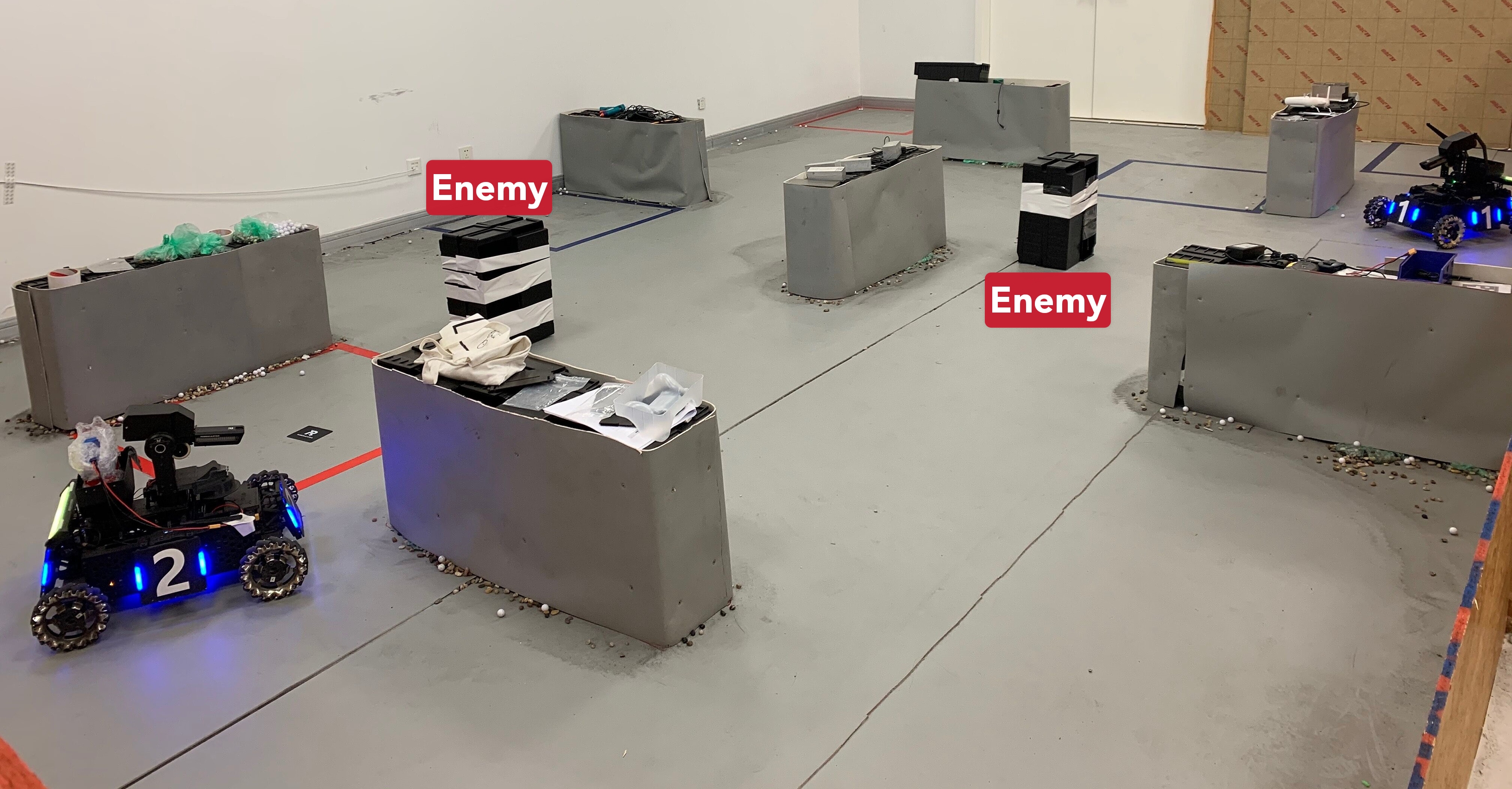}
    \caption{The arena used for experiments. As enemies don't change position during iterations, we use two plastic boxes (in black, at the figure) to emulate their positioning, forcing our robots to use LiDAR sensors to localize them.}
    \label{fig:experiment}
\end{figure}

In juxtaposition, the opposing trend among RL algorithms consists of model-based approaches, which devise controllers from a predictive transition model of an environment. Model-based methods are capable of fast learning due to their sample efficiency and, as a result, they can be directly applied to real-world robotics experiments and skip the reality gap. These methods consist of the learning of a probabilistic or Bayesian transition model, which translates into a higher sample efficiency \cite{PILCO2011, DeepPILCO2016, wangshu-robio, PolSearch2019}. For example, Deep PILCO \cite{DeepPILCO2016} is a typical probabilistic model-based RL algorithm that relies on a Bayesian neural network (BNN) transition model. Although computationally expensive, it advanced the DQL algorithms in terms of the number of trials by at least an order of magnitude on the cart-pole swing benchmark task.

Deep PILCO was already applied to a number of robotic tasks, where \cite{DPcode2018} improved Deep PILCO by using random numbers and gradient clipping, and applied it for learning swimming controllers for a simulated 6-legged autonomous underwater vehicle. In \cite{kahn2017uncertainty}, the authors learned
the specific task of a quad-rotor and an RC car navigating an unknown environment while avoiding collisions using Deep PILCO with bootstrap \cite{bootstrap1982jackknife}.
The advantages of Deep PILCO in learning speed have been proven on simulations and single-robot experiments. In this paper, we will further demonstrate its potential in applications within a noisy real-world environment in the context of a multi-agent competitive game.

Here, we apply Deep Q-Learning (DQL) and Deep PILCO on simulations and real-world experiments of a robot combat decision making problem. This problem consists of the control of a robot positioning itself in an arena to shoot at enemies. We compare the two aforementioned algorithms on a Gazebo simulation of those robots and also on real-world experiments. Our results show that Deep PILCO was superior to DQL in the speed of convergence and in the quality of its best policy on both simulations and experiments. More importantly, the real-world implementation of the Deep Bayesian algorithm found the optimal solution in 20 minutes, which is faster than the real-time deployment of the Deep Q Learning algorithm in both real-world and simulation.

\begin{figure}[t]
    \centering
    \includegraphics[width=0.8\columnwidth]{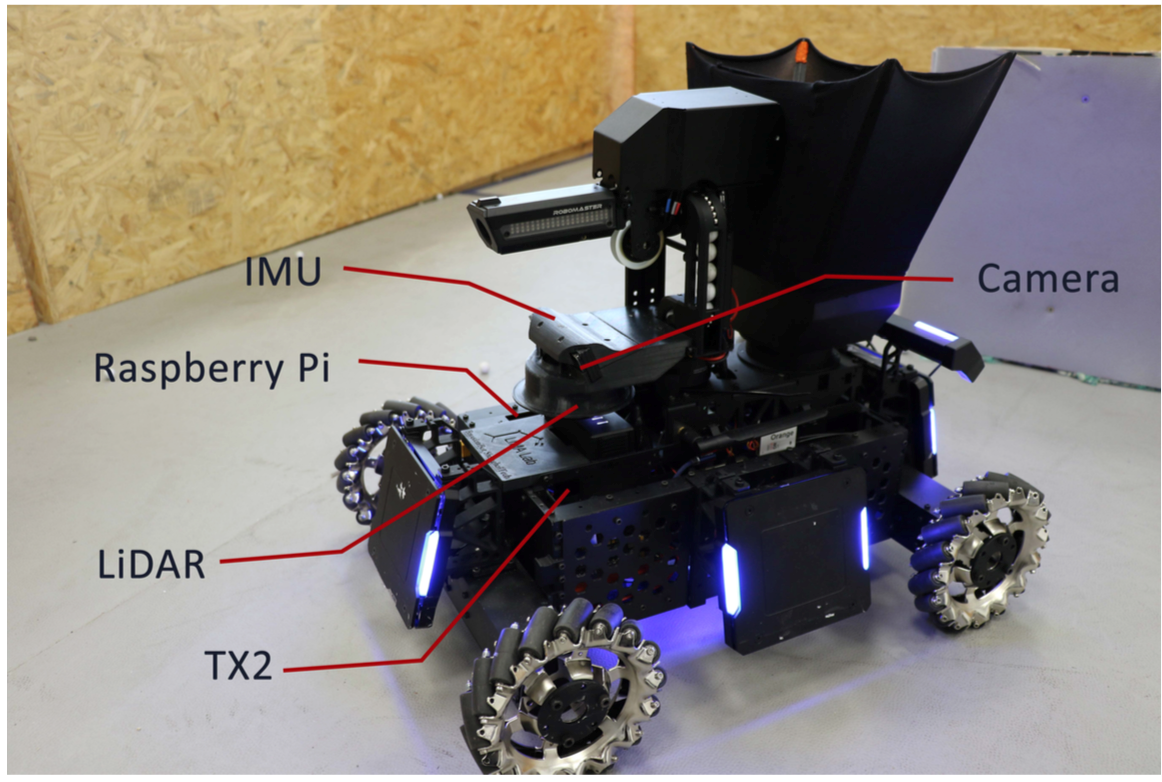}
    \caption{Hardware of the adopted robot. The robot is capable of recognizing the enemy through a combination of a LiDAR and a camera, both sensors sampled by a TX2 and a Raspberry Pi.}
    \label{fig:hardware}
\end{figure}

\section{Problem Definition}
The decision-making problem comes from a 2v2 robot competition called ICRA-DJI RoboMaster AI Challenge. During the match, the most important decision for the robot to make is where to go for firing projectiles based on the current situation. From a robot's view, the number of enemies within sight has three possibilities - zero, one, and two. 

We can develop an appropriate reward mechanism to achieve an optimal situation. We formulate the reward function as:
\begin{equation}
R(s)= \left\{
\begin{aligned}
    0, \quad & s[N_{E}] \ne n \\
    1, \quad & s[N_{E}] = n
\end{aligned}
\right.
\end{equation}
Here,  $n$ is the target number of visible enemy robots. In this experiment, discovering one enemy does not necessarily form the sub-task of finding both enemies, so the reward is not defined to be proportional to the number of visible enemies.

As an Markov Decision Process, the state space contains all the possible positions the robots can locate at in the discretized map and the number of detected enemy robots, where the state is $(p_{M}, p_{E_{n}}, N_{E})$ with $M$ denoting the robot itself and $E_{n}$ denoting the enemy robots. The action $(p_{G})$ is the next goal position for the controlled robot, where the action space contains four nearest neighbours of the current position $p$.

\section{Materials and Methods}
\subsection{Experimental Design}
We conducted simulations and real-world experiments on a 5m x 8m map (Fig. \ref{fig:experiment}) with robots shown in Fig. \ref{fig:hardware} and obstacles on it. During our training we emulate enemy positions with plastic boxes. 

We initially tuned the hyper-parameters of DQL and Deep PILCO on a Gazebo simulation \cite{koenig2006gazebo} to reduce the expensive cost of running real robots. Both algorithms perform calculations in a computer with 12-core Intel i7 CPU. The robot (agent) uses a LiDAR sensor and IMU to detect enemies and to improve its own self-localization, and those signals are processed by the main computer. In our simulated environment those signals are also emulated from the simulated robot, to emulate noise and disturbance generated by the sensor while detecting the enemy position. The simulation clock is capable of running 2.4 times faster than real-time experiments, and both simulated and real implementations have their speed capped by the processing time.

We run simulations and experiments in two cases: The first case is 1v1 design, which means that there is only one robot against one enemy robot, while the second case is 1v2 design, as there is one robot against two enemy robots.

\begin{figure}[b]
    \centering
    \includegraphics[width=0.8\columnwidth]{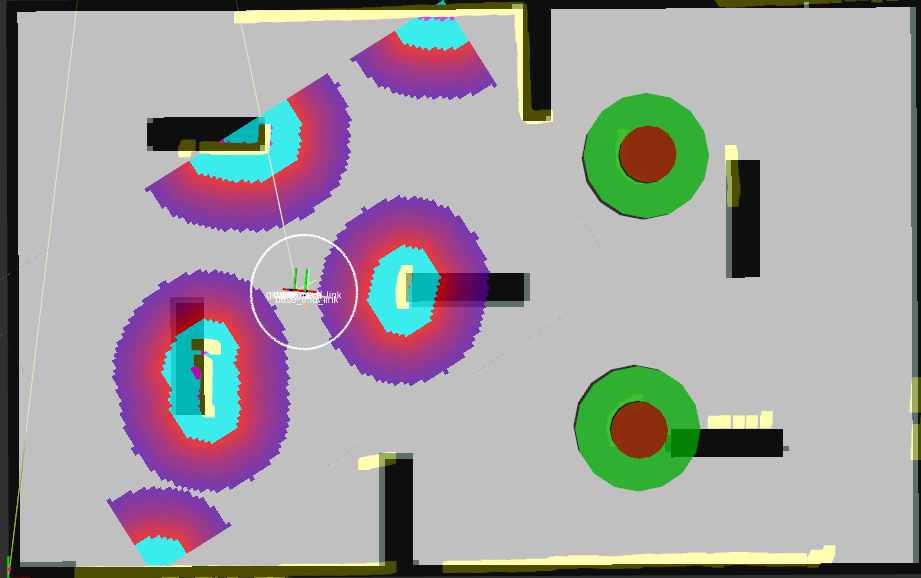}
    \caption{The visualization of the LiDAR-based enemy detection algorithm. The position of the plastic boxes (enemies) are shown in the two green circles. The navigation stack in ROS depicts the contour of the obstacles in yellow, and displays the local costmap in blue, red and purple.}
    \label{fig:rviz}
\end{figure}

\subsection{Deep Q-Learning}
DQL is a variant of the Q-Learning \cite{watkins1992qlearning} algorithm, which takes a deep neural network as a Q-value function approximator where samples are generated from the experience replay buffer. 
\begin{equation}
    Q^*(s,a) = E_{s'}[r+\gamma \max_{a'}Q^*(s',a')|s,a].
\end{equation}
Note that DQL is model-free: it solves the RL task directly using samples from the emulator, without explicitly constructing an estimate of the emulator \cite{DQN2016}. Instead of updating the policy once after an episode in the model-based algorithm PILCO \cite{PILCO2011}, DQL updates the Q-network with samples from the replay buffer every step.

We implemented the DQL algorithm using the Tianshou library \cite{tianshou}. As for the model architecture, the input to the Q-network is a state vector. The two hidden layers consist of 128 neurons for simulations, 16 neurons for experiments, activated by ReLU function. The output layer is a fully-connected linear layer with a single action output. The policy during training is $\epsilon$-greedy at $\epsilon=0.1$. The learning rate is 0.001, and the discount factor is 0.9. The size of the replay buffer is 20000.

\begin{figure}[b]
    \centering
    \includegraphics[width=0.8\columnwidth]{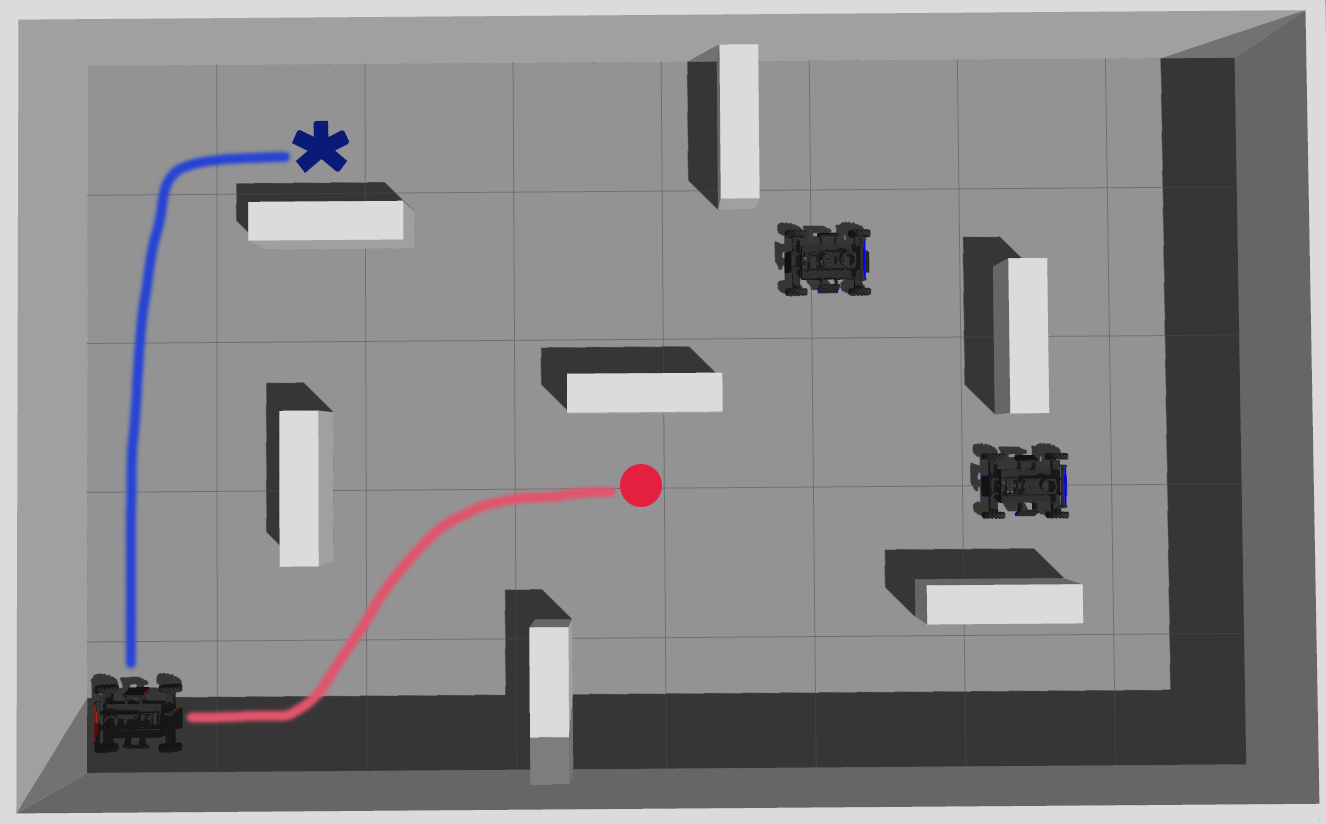}
    \caption{The simulated arena in GAZEBO and the optimal paths generated by DQL and Deep PILCO. The blue path is learned by DQL. The red path is learned by Deep PILCO.}
    \label{fig:gazebo-screen}
\end{figure}

\subsection{Deep PILCO}
Compared to model-free deep RL algorithms, model-based RL allows higher sample efficiency, which can be further improved with a probabilistic transition model. Deep PILCO is a prominent example which utilizes a Bayesian neural network (BNN) \cite{BNN1992} to estimate the transition model \cite{DeepPILCO2016, DPcode2018}.

We implement Deep PILCO in an episodic way so that the algorithm updates the policy after every episode based on the episodic rewards. The episodic reward is the sum of iteration rewards. Each episode consists of 10 iterations. During one iteration, the robot moves from the current position to the goal position given by the action along the planned path. The code is a modified version of an open-source implementation of Deep PILCO algorithm \cite{DPcode2018}.

\begin{figure}[b]
    \centering
    \includegraphics[width=\columnwidth]{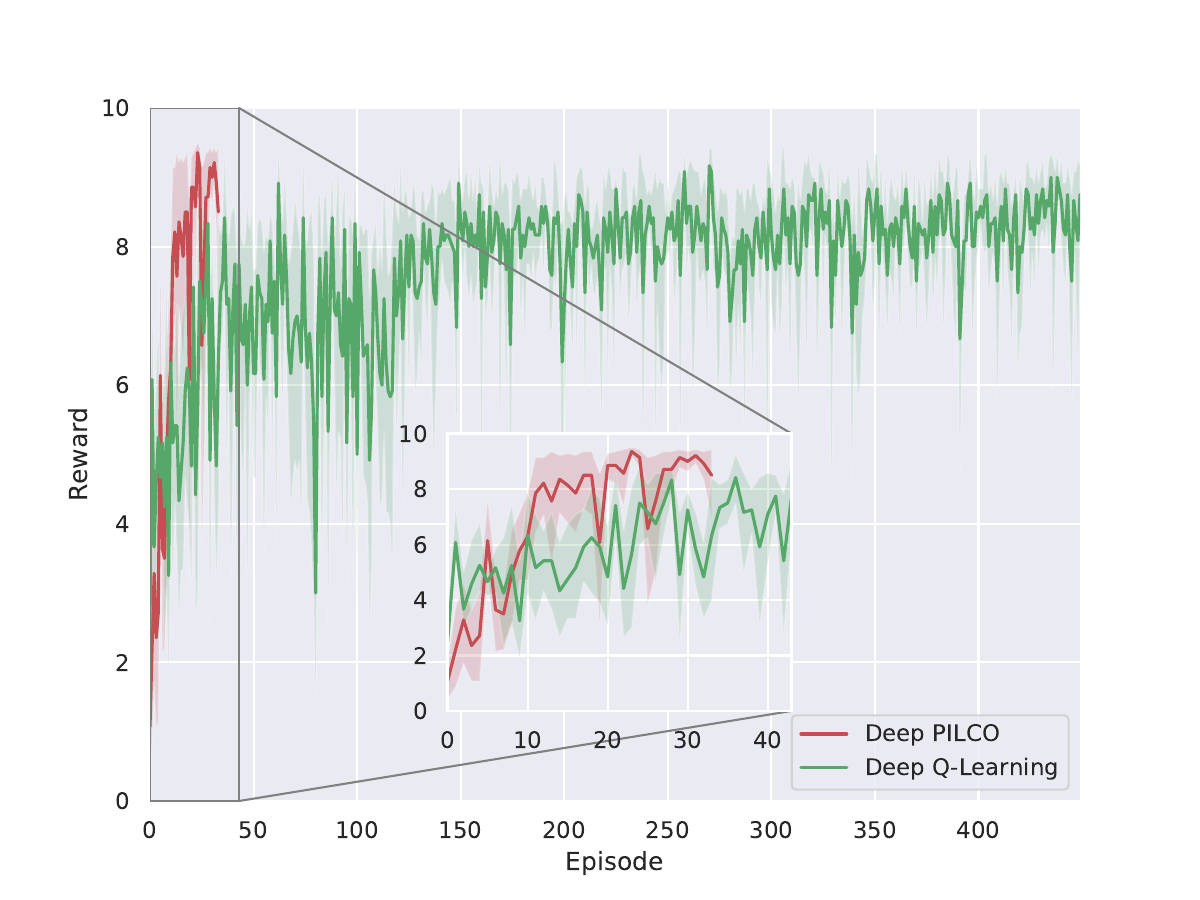}
    \caption{Results for the 1v2 case task of DQL and Deep PILCO in the GAZEBO simulator (seven replicates). The shaded regions show the 95\% confidence intervals around the mean line in the middle.}
    \label{fig:gazebo}
\end{figure}

\begin{figure*}[tb]
    \centering
    \subfloat[1v1 case]{\includegraphics[width=\textwidth]{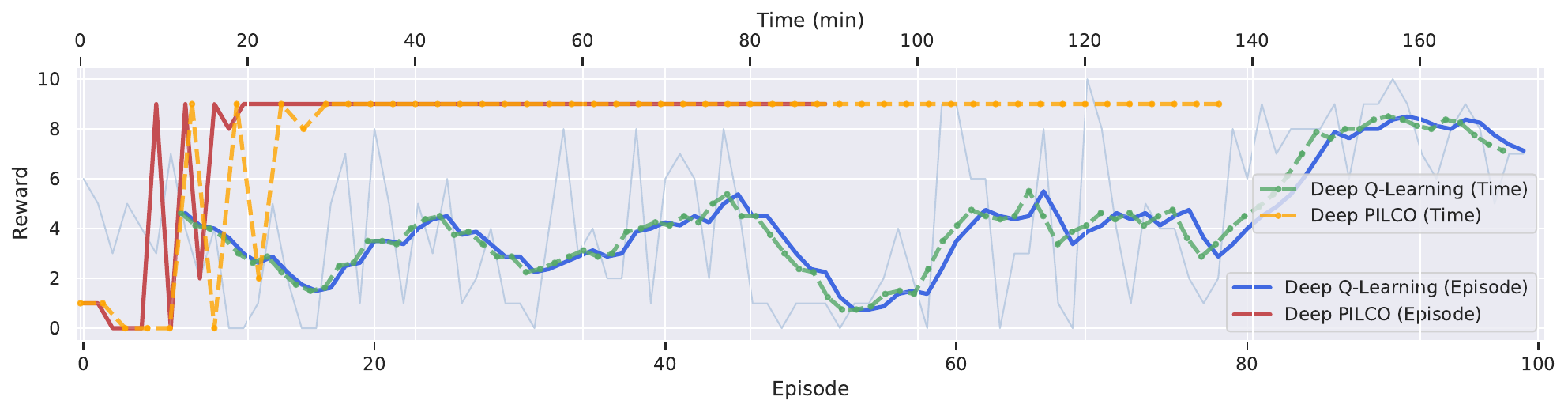}\label{fig:res-a}}
    \hfil
    \subfloat[1v2 case]{\includegraphics[width=\textwidth]{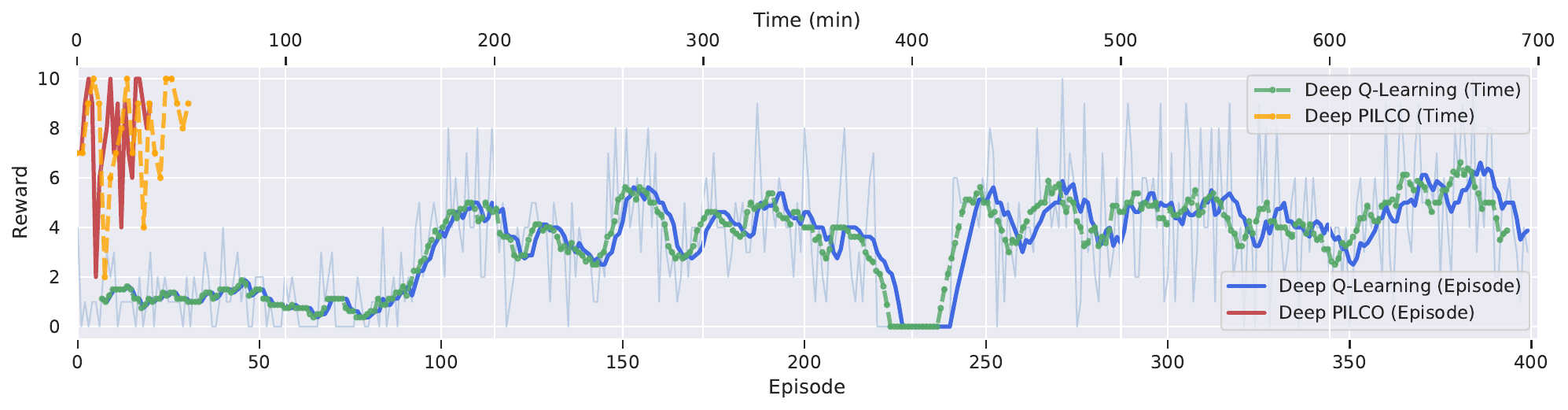}\label{fig:res-b}}
    \caption{Training rewards versus episodes (solid lines) and versus wall-time (dashed lines) of DQL and Deep PILCO. For DQL vs. episode, the rolling mean rewards are drawn to reduce noise and show convergence. For both methods, each episode contains ten iterations to execute the policy, with each iteration spending about ten seconds. As for the computation time, Deep PILCO takes approximately 60 seconds per episode, while DQL takes three seconds per episode. Combining execution time and computation time, the total training time is 160 seconds per episode for Deep PILCO, and 103 seconds per episode for DQL.
    (a) Results for 1v1 case. The red curve vanished earlier since Deep PILCO converged to the optimal reward within fewer training episodes. (b) Results for 1v2 case.}
    \label{fig:result}
\end{figure*}

\subsection{LiDAR-based Enemy Detection Function}
We detect enemies with a 2-D LiDAR sensor by filtering out the known obstacles in the given map. Since we already know the map, we know where the walls and obstacles are. If the center of a detected obstacle is inside a wall, we filter out this circle \cite{YiZheng2019}. Otherwise, it will be recognized to be an enemy. In the real world experiments, we replace the two enemies by two black boxes to save usages of real robots. A screenshot taken during the algorithm running is presented in Fig. \ref{fig:rviz}.

\section{Results} \label{Results}
\subsection{Simulation Results}

We simulate the arena and the robots with a one-to-one ratio within GAZEBO, as shown in Fig. \ref{fig:gazebo-screen}. The simulation time is 2.4 times faster than in real-time. We focus on simulating the combat environment with a 1v2 case. Fig. \ref{fig:gazebo} presents the simulation results of DQL and Deep PILCO. Although both of them converge to an optimal policy, Deep PILCO achieves a higher best reward than DQL, within fewer episodes.

\begin{figure*}[htb]
    \centering
    \includegraphics[width=\textwidth]{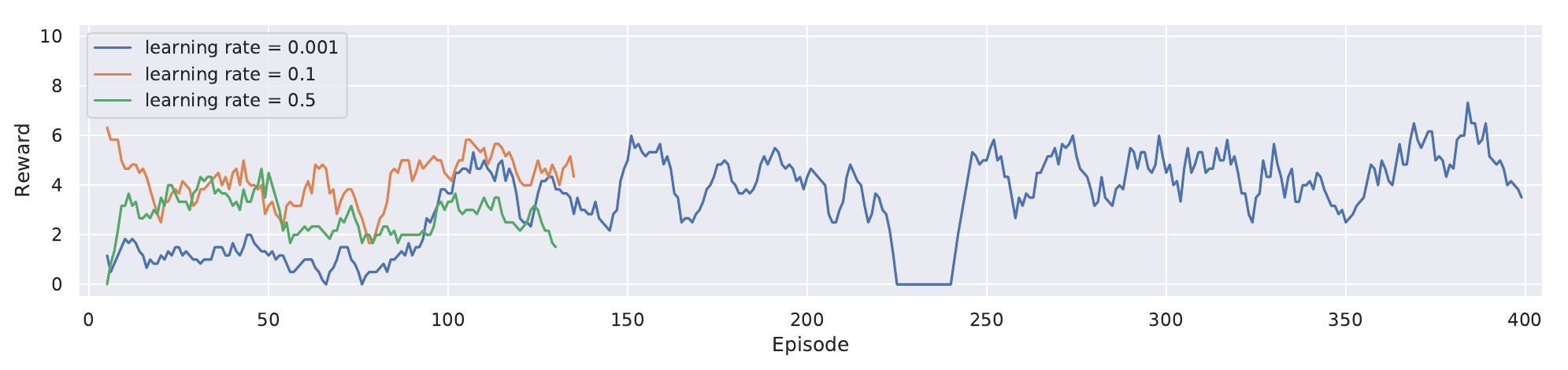}
    \caption{DQL training results for three learning rates at the 1v2 case. Empirically, learning rates with large values hinder the convergence in DQL experiments. With that in mind we ran more trials with the smallest learning rate 0.001. Nonetheless, all three experiments failed to achieve a high reward.}
    \label{fig:alpha}
\end{figure*}

\subsection{Experimental Results}

In the real world experiments, we implement the 1v1 case and 1v2 case individually. Here, we analyze the results from two perspectives, number of training episodes and length of computation time.

We first compare the episodic rewards of DQL and Deep PILCO. In order to reveal the learning trend, we also plot the rolling mean rewards of six neighboring episodes for DQL. For the 1v1 case, both algorithms learned optimal solutions after training. In Fig. \ref{fig:res-a}, we can see that Deep PILCO found the solution within 11 episodes, much fewer than DQL, which took around 90 episodes. Furthermore, the result of Deep PILCO remained near its maximum even after the optimal moment, while the result of DQL was more unsteady.

For the 1v2 case, the results of both algorithms fluctuated more than in the 1v1 case, as shown in Fig. \ref{fig:res-b}. While the performance of Deep PILCO kept a similar number of episodes when changing from the 1v1 case to the 1v2 case, DQL failed to converge to an optimal solution after 400 training episodes.

Considering the expensive training cost of real-world experiments, we decided to stop the experiment after 400 episodes. To eliminate the impact of the hyper-parameters, we changed the learning rate parameter of the DQL algorithm and re-ran the experiments, but DQL was still unable to find a stable optimal solution, as we can see in Fig. \ref{fig:alpha}. Higher learning rates are likely to lead to a performance breakdown \cite{Hyper2019}.
For this reason, we stopped the experiments earlier for the two higher learning rates.

\begin{figure}
    \centering
    \includegraphics[width=\columnwidth]{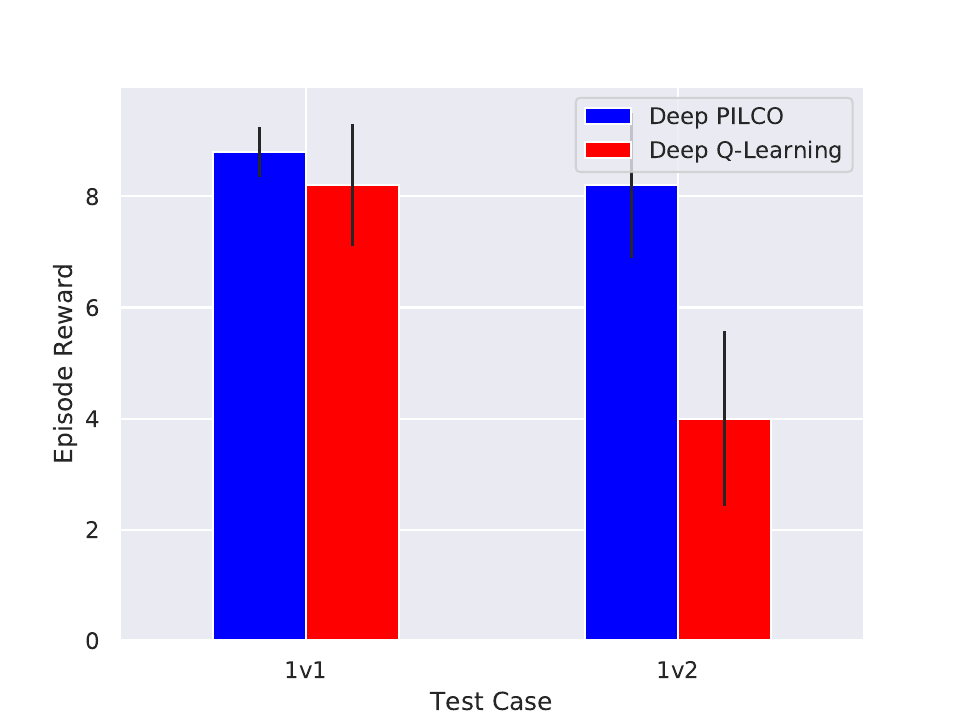}
    \caption{Statistic results of real-world Deep PILCO and Deep Q-Learning methods for two test cases. One-episode tests were run for ten times for each case. The histogram shows the average episodic reward value for different metrics, while error bars show the standard deviation.}
    \label{fig:test}
\end{figure}

With regards to the computation time, fewer episodes are not necessarily equivalent to a shorter training time, since each episode costs a different wall-time for DQL and Deep PILCO. Consequently, we plot the same results in the reward-versus-minute figures to compare the converging speed of these two algorithms. Fig. \ref{fig:result} shows that in both the 1v1 case and 1v2 case, Deep PILCO achieved the optimal point much faster than DQL. Deep PILCO required less than one hour to find the optimal solution.

\begin{figure}
    \centering
    \includegraphics[width=0.8\columnwidth]{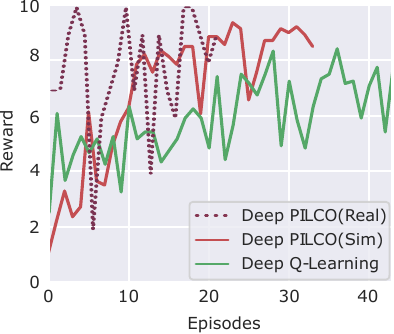}
    \caption{Comparison between learning with Deep PILCO in simulation and real-world on a 1v2 case, with Deep Q-learning also added for reference. Straight out of the rollout phase the real-world performance of Deep PILCO apparently outperforms simulation results, but these might be the result of a ``luckier" rollout phase. Nonetheless, as the transferability from simulated results to the real-world is often poor, the real-world learning has the advantage.}
    \label{fig:sim2real}
\end{figure}

After the training procedure, we evaluate the optimal policies of DQL and Deep PILCO and compare the average episodic rewards. Each testing experiment ran for only one episode. For each case, we executed ten tests and computed the average episodic rewards. The histogram in Fig. \ref{fig:test} demonstrates that in both 1v1 and 1v2 cases, Deep PILCO scored higher rewards than DQL.

We further analyze the training efficiency of real-world experiments and simulations together by comparing the number of iterations and the minimal total training time required for each method to exceed the test reward threshold at the 1v2 case. The results are shown in Fig. \ref{fig:sim2real}. Deep PILCO in the simulation beats all the other cases. Nonetheless, in the real-world experiment, it requires nearly the same amount of wall-time, showing the potential of Deep PILCO to bypass the reality gap. While DQL fails to learn a solution to pass the test within acceptable time in the real world, it is unanticipated that even in the simulation, DQL still requires several times longer training time than real-world Deep PILCO.

Fig. \ref{fig:snapshot} displays snapshots of our experiments with Deep PILCO for the 1v2 case. The four pictures show different phases of the found optimal policy. We can see that our robot started from the initial state, where it saw none of the enemy robots, and finally navigated to an optimal position where it could see two enemy robots at the same time. In the rest of the episode, it stayed at the optimal position in order to achieve the highest episodic reward.

\section{Discussion}

\subsection{Superiority of Deep Bayesian RL over Deep RL}
The results of the simulations and experiments indicated that Deep Bayesian RL surpassed Deep RL in both learning efficiency and learning speed. This corroborates the findings of previous works in \cite{PILCO2011, DeepPILCO2016} that explicitly incorporated model uncertainty and temporal correlation into planning and control, enhancing the sample efficiency and scalability. Although DQL required much shorter computation time than Deep PILCO for each iteration, the learning efficiency of the latter compensated that cost.
According to Moore's law, the number of transistors on a chip should double in each technology generation. We presume that Deep Bayesian RL algorithms will be much more superior to Deep RL in terms of learning speed, on the basis of predictable advances in computation hardware.

Note that while DQL didn't perform well in the 1v2 case, with the original hyper-parameters fine-tuned for other applications in \cite{DQN2016}, Deep PILCO was still capable of learning the best policy, without the extra effort of modifying hyper-parameters, which were chosen for the basic cart-pole swing up experiment in \cite{DPcode2018}. This finding suggests Deep PILCO demands less work than DQL to fine-tune hyper-parameters every time we apply it on a new kind of task. This property makes Deep PILCO more flexible for various applications.

It is also worthy of mention the superiority of real-world experiments with Deep PILCO in comparison to simulated results from DQL. Although simulations are widely accepted as a faster media than the real world, the efficiency behind the aforementioned Deep Bayesian algorithm was sufficient to overcome that problem.

\subsection{Effect of Random Rollouts in Deep PILCO}
Observing the first few episodes of the training curves of Deep PILCO in Fig. \ref{fig:result}, we noticed that the initial reward of the 1v2 case was already much higher than that of the 1v1 case. This implies that the initial random rollouts in the 1v2 case explored more beneficial trajectories so that the first learned dynamics model was already expressive enough to achieve a high reward. Nonetheless, there is no guarantee such that a better initialization leads to faster learning.
Likewise, increasing the number of initial random rollouts also did not enhance the performance evidently during our experiments, which is consistent with the finding of \cite{MbMf2018}.
In their work, which combines model predictive control (MPC) and reinforcement learning, they evaluated various design decisions and found that low-data initialization runs were able to reach a high final performance level as well due to the reinforcement data aggregation.

\subsection{DQL's Lack of Convergence}
As for the lack of convergence issue of DQL in 1v2 case experiments, the basic function approximation idea of Q-Learning could be a major factor, if not the only one, causing convergence failures. Recently, \cite{Diagnosis2019} introduced a unit testing framework to investigate the effect of function approximation on convergence. They surprisingly found that the function approximation rarely caused divergence in Q-Learning algorithms, but only when the representational capacity of the function approximator is high. Namely, the network architecture should not be too small.
\begin{figure}[t]
    \centering
    \includegraphics[width=\columnwidth]{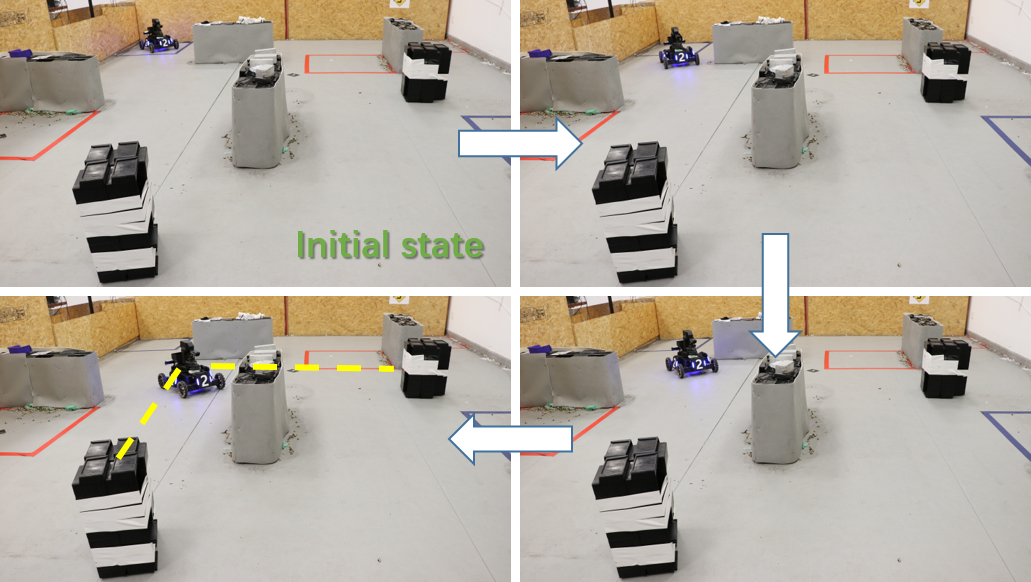}
    \caption{Snapshots of the real-world experiment for the 1v2 case. After about 15 training episodes, the robot found the optimal position to see the two enemies simultaneously. During the episode that the reward is the highest, the robot started from the initial position and then navigated to the optimal place at the end of the first iteration. The robot stayed at the optimal place during the rest of the episode to reach a maximal reward.}
    \label{fig:snapshot}
\end{figure}

We suggest more focus on improvement in the computation time of Deep Bayesian algorithms in future work. The algorithm can be faster when multiple cores are available if we replace the gradient-based optimization algorithm with a parallel, black-box optimizer. \cite{wangshu-robio,chunyanICAR}

\section{Conclusions}
We proposed a new application of Deep PILCO on a real-world multi-robot combat game. We further compared this Deep Bayesian RL algorithm with the Deep Learning-based RL algorithm, DQL. Our results showed that Deep PILCO significantly outperforms Deep Q-Learning in learning speed and scalability, and the real-world learning rate is even faster than the learning rate shown by DQL on simulations. We conclude that sample-efficient Deep Bayesian learning algorithms have great prospects on competitive games in the real world, as opposed to being limited to simulated applications.

\bibliographystyle{ieeetran}
\bibliography{main}

\end{document}